\documentclass[11pt]{article}

\usepackage[utf8]{inputenc}
\usepackage[T1]{fontenc}
\usepackage[margin=1in]{geometry}
\usepackage{amsmath,amssymb}
\usepackage{enumitem}
\usepackage{booktabs}
\usepackage{array}
\usepackage{tabularx}
\usepackage{titlesec}
\usepackage[colorlinks=true]{hyperref}
\usepackage[numbers,sort&compress]{natbib}
\usepackage{xcolor}
\definecolor{refblue}{RGB}{31, 78, 155}
\hypersetup{linkcolor=refblue, citecolor=refblue, urlcolor=refblue}
\usepackage{graphicx}
\usepackage[most]{tcolorbox}
\usepackage{float}
\usepackage{algorithm}
\usepackage{algpseudocode}
\algtext*{EndIf}
\algtext*{EndWhile}
\newcommand{\doi}[1]{doi: \href{https://doi.org/#1}{\nolinkurl{#1}}}

\titleformat{\section}{\large\bfseries}{\thesection}{0.6em}{}
\titleformat{\subsection}{\normalsize\bfseries}{\thesubsection}{0.6em}{}
\setlist{itemsep=2pt,topsep=3pt}
\begin{document}
\title{\textbf{Efficiency Matters in Autonomous Research}}
\author{
    Haiqian Yang \\
    \small \texttt{hqyang@mit.edu}
    \and
    Yuan Cao \\
    \small \texttt{realcaoyuan@gmail.com}
}
\date{\today}
\maketitle

\begin{abstract}
\noindent
AI-driven autonomous research (AR) systems are becoming increasingly effective across a broad range of tasks. Their performance, however, is still evaluated primarily by the quality of the final outcome. In this paper, we argue that the efficiency of the solution-search process is an equally important but often overlooked dimension of performance. A strong AR system should not only produce high-quality results, but also reach them with as small a budget as possible.
Search efficiency will become increasingly important as AR expands from domains with inexpensive verification, such as mathematics and coding, to real-world scientific settings in which solution evaluation may require costly physical experiments. To capture this dimension, we propose evaluating AR systems using the area under the curve (AUC) of the Pareto frontier, alongside final outcome quality.
We compare several families of search algorithms, including hill climbing, beam search, tree search, and evolutionary search, across twelve systems-optimization tasks. We find that no single search structure is consistently the most efficient. We also show that search efficiency and final outcome quality are distinct performance dimensions: a method that eventually achieves the best result may nevertheless improve slowly and consume substantially more evaluation budget before reaching that result. Because the most effective search policy is generally unknown in advance, we introduce an adaptive procedure called \texttt{fluid} search, which uses a portfolio bandit to dynamically allocate a fixed evaluation budget across a forest of search processes. Across the evaluated tasks, \texttt{fluid} search achieves the highest overall search efficiency, closely matching the performance of a per-task oracle that is given the best search structure for each task in advance.

\end{abstract}

\begin{figure*}[h]
    \centering
    \includegraphics[width=0.95\linewidth]{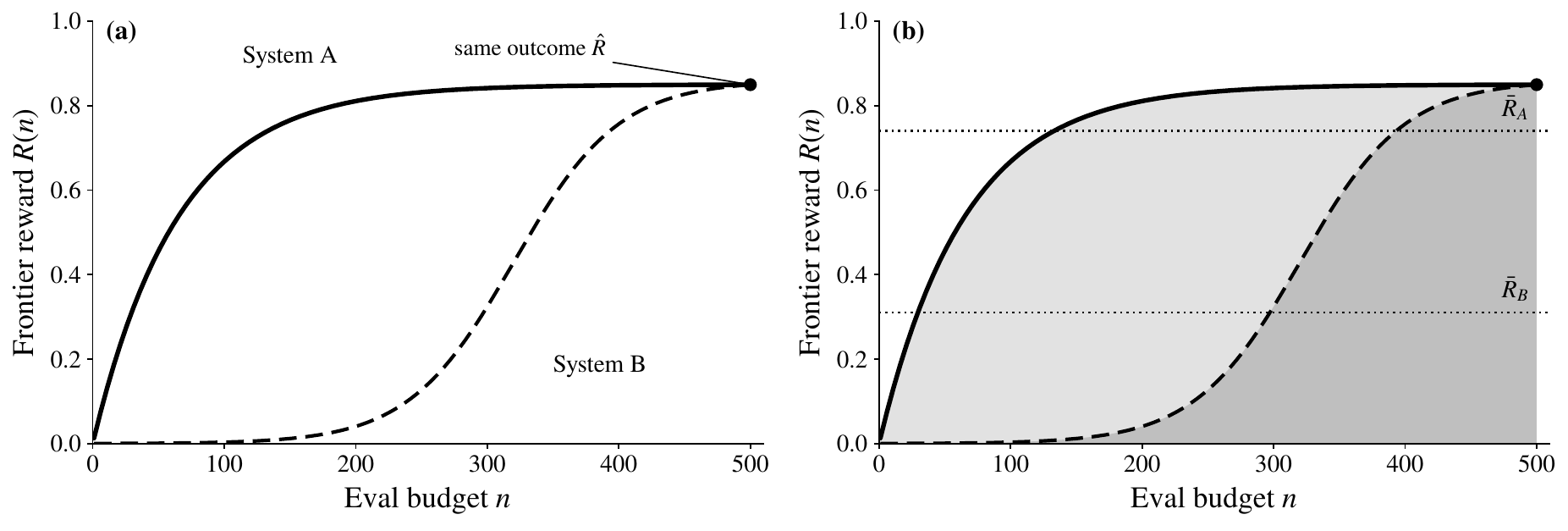}
    \caption{\scriptsize
    The two scalar summaries of a Pareto frontier of research quality, on two hypothetical systems that end at the same peak. (left) Frontier reward \(R(n)\): the systems reach the same final outcome at \(\hat{R} = R(N)\), but with the same budget System A achieves superior solutions earlier than B. (right) The AUC reward \(\bar{R}\), the mean height of the area under each frontier (dashed lines), separates them: \(\bar{R}_A > \bar{R}_B\) because System A conducts research more efficiently.}
    \label{fig:illustration-metrics}
\end{figure*}

\clearpage
\section{Introduction}

AI-driven autonomous research (AR) systems are becoming an increasingly promising approach to solving difficult problems across mathematics, algorithm design, software engineering, materials discovery, and other scientific and engineering domains~\citep{kong2026aiautoresearchroadmap}. By repeatedly proposing hypotheses or candidate solutions, evaluating them, and using the resulting feedback to guide subsequent decisions, these systems can conduct increasingly long and complex research processes with limited human intervention. Existing AR systems are typically evaluated primarily by the quality of the best final solution they discover. This outcome-based evaluation is natural but incomplete. A capable AR system should not only find strong solutions, but find them within a reasonable budget. A system that eventually discovers an excellent solution only after years of computation or billions of dollars in experiments would be of limited practical value, regardless of the quality of its final result.

We therefore argue that AR should be evaluated along two distinct dimensions: \emph{final outcome quality} and \emph{research efficiency}. Evaluating only the best solution found conflates these two dimensions and ignores the search trajectory by which that solution was reached. Accordingly, we propose evaluating the full performance curve over the available budget, using the area under the curve (AUC) of the Pareto frontier as a measure of research efficiency. A smart AR system should not only find outstanding solutions, but also develop strong intuitions and find them early. This broader view will become increasingly important as AR moves beyond cheaply verifiable tasks such as mathematics, coding, and algorithm optimization toward discovery in the physical world. In many scientific and engineering settings, evaluating a candidate may require laboratory experiments, physical fabrication, clinical studies, or other costly interventions. The structure of the problem may also be only partially known or incomplete. Under these conditions, inefficient exploration is not merely an inconvenience, it can make such an AR system practically unusable.

The principle that a search procedure should be judged by its efficiency rather than only by its endpoint is well established in optimization. Anytime performance and the AUC over a fixed budget are standard evaluation objectives in areas such as AutoML~\cite{feurer2015efficient} and molecular design~\cite{zilberstein1996anytime, gao2022pmo}. The budget is defined by the dominant cost in the research loop.\footnote{When evaluating a candidate dominates the total cost, as in research requiring physical experiments or expensive simulations, the budget is naturally measured by the number of evaluations. When evaluation is inexpensive, the relevant budget may instead be the number of search steps or the model computation used to generate candidate edits. In this work, we focus on evaluation cost and report the budget as the number of evaluations.} To illustrate how different search policies affect AR research efficiency, we compare four canonical search families spanning local refinement, explicit breadth, tree-based exploration, and population-based search (greedy hill climbing, beam search, Monte Carlo tree search, and an evolutionary strategy) on 12 tasks from AutoLab~\cite{autolab}. Each method scaffolds the same LLM proposer, which generates a single candidate solution edit conditioned on the preceding evaluation history. The proposer, evaluator, and total budget are held fixed, only the search structure differs.

\begin{tcolorbox}[
  colback=orange!6,
  colframe=orange!30!brown,
  boxrule=0.6pt,
  arc=1mm
]
\textbf{Contributions}:

\begin{enumerate}
    \item \textbf{No single search structure is uniformly the most efficient.} Each structure embodies assumptions about the underlying optimization landscape and performs well when those assumptions are aligned with the task. Beam search is effective when a fast-deepening early frontier identifies promising regions whose advantages compound over time. Evolutionary search is effective when maintaining a population protects against frequent catastrophic edits. Greedy search works well when local refinements are reliable, and tree-based search is useful when promising solutions require exploring multiple branching trajectories. Because no single set of assumptions fits every task, the best fixed search policy is inherently task dependent.

    \item \textbf{Efficiency and final solution quality are empirically distinct axes.} A method that improves slowly but achieves a late breakthrough may attain the highest final score while performing poorly in AUC. Conversely, a method that produces rapid and reliable early gains may be the most useful under realistic budgets even if another method eventually surpasses it. Endpoint evaluation alone therefore provides an incomplete, and sometimes misleading, picture of AR performance.

    \item \textbf{We introduce \texttt{fluid}, a budget-adaptive search policy that avoids committing to a search structure in advance.}
    Search trajectories provide online evidence about how well the searching process is doing, including each concurrent chain's improvement rate, crash frequency, and incumbent quality.
    \texttt{fluid} combines multi-start local search with bandit-based allocation for LLM-driven AR, dynamically directing a fixed evaluation budget toward the chains showing the greatest progress.
\end{enumerate}
\end{tcolorbox}

Given a fixed evaluation budget, \texttt{fluid} reallocates evaluations among its parallel chains, concentrating resources on those that continue to improve while preserving sufficient exploration to avoid permanently starving the others. Across twelve tasks, \texttt{fluid} is the most efficient single policy evaluated, nearly matching a per-task oracle that is given the best search structure for each task in advance. Moreover, \texttt{fluid} matches the fastest fixed policy over the opening budget and accumulates its advantage over the remainder of the run. Its near-oracle performance suggests that the signals needed to allocate the budget well are already present in the trajectory. Consequently, the cost of not knowing the most suitable search structure in advance can be made negligible.

Together, these results support a broader evaluation paradigm for AR. Progress should be measured not only by the best discovery a system eventually produces, but also by how efficiently it reaches useful discoveries along the way. As AR systems are deployed in increasingly expensive and consequential domains, this distinction will become essential for both scientific evaluation and practical system design.~\footnote{Code available at: \url{https://github.com/HaiqianYang-MechE/AREK}.}

\section{Related work}

\noindent\textbf{Search over candidate solutions.}\enspace
Searching over a space of candidate solutions is a classical approach in problem solving and intelligence modeling. Game-playing systems search move trees~\cite{deepblue, alphago}, neural architecture search explores network designs~\cite{nas}, and evolutionary computation has long been used to optimize programs~\cite{koza}. The policy families studied here are drawn directly from this classical repertoire. What changes in AR is the proposal operator. Whereas earlier methods relied on a hand-designed move generator, sampler, or mutation operator, an LLM now proposes the next state conditioned on the task and the search history. The underlying search structures and questions of efficiency therefore remain relevant, even though the mechanism for generating candidates has changed. Whether the classical ordering of these search strategies persists under an LLM-based proposal operator is an empirical question that we investigate.

\noindent\textbf{Autonomous research and code agents.}\enspace
The repeated use of an LLM to modify an artifact and improve a verifier-provided score forms the core loop of program-search systems such as FunSearch~\cite{funsearch} and AlphaEvolve~\cite{alphaevolve}, as well as ``AI scientist'' agents that augment this loop with planning, tool use, and memory~\cite{weng2023agent,react,ren2026selfimprovementsmodernagenticsystems}. We deliberately isolate the search component of this broader stack: each proposal uses a single LLM completion, without a multi-turn agentic loop, so that search structure is the primary experimental variable. Existing benchmarks typically evaluate such systems by an endpoint metric, such as resolved-issue rate, medal rate, or the best score attained within a fixed budget~\cite{swebench, mlebench, kernelbench, autolab}. RE-Bench~\cite{rebench} additionally tracks performance as a function of time, but uses these trajectories descriptively to compare agents with human experts at fixed checkpoints rather than treating the full trajectory as the optimization objective. Closest to our first research question, Toledo et al.~\cite{toledo2025aira} compare greedy, tree-based, and evolutionary search on MLE-bench, while AIDE~\cite{aide} applies a single tree-search procedure to code optimization. Both evaluate methods primarily by their final outcomes. In contrast, we include beam search as a peer strategy and compare all search structures by their efficiency over the full evaluation budget.

\noindent\textbf{Search efficiency and adaptive allocation.}\enspace
The anytime perspective adopted in this work---evaluating a search procedure by its entire progress curve rather than only its endpoint---is well established in sample-efficient optimization~\cite{zilberstein1996anytime, gao2022pmo}. Our contribution is to bring this perspective to AR and examine its implications for search design. In LLM-based inference, test-time search has commonly used a fixed structure, such as a search tree or an iterative refinement chain~\cite{tot, lats, reflexion, selfrefine}. Recent work further shows that performance depends not only on the total test-time budget but also on how that budget is allocated~\cite{snell2024scaling}.

Our proposed \texttt{fluid} search procedure views a collection of restarting hill-climbing chains~\cite{multistart, ils, grasp} as an algorithm portfolio~\cite{rice1976algselection, gomes2001portfolios} and allocates evaluations using a UCB index~\cite{auer2002ucb}. As a result, a fixed evaluation budget is directed according to observed progress rather than a predetermined restart schedule~\cite{lubyrestart}. Adaptive budget allocation is well established in settings ranging from Hyperband for AutoML~\cite{hyperband} to bandit-based allocation across acquisition strategies, tree branches, parallel evolutionary runs, and LLM choices~\cite{brochu2010gphedge, abmcts, base, shinka}. \texttt{fluid} applies this principle to a portfolio of LLM-driven hill-climbing chains, using anytime performance as the central objective.


\begin{figure*}[h]
    \centering
    \includegraphics[width=\linewidth]{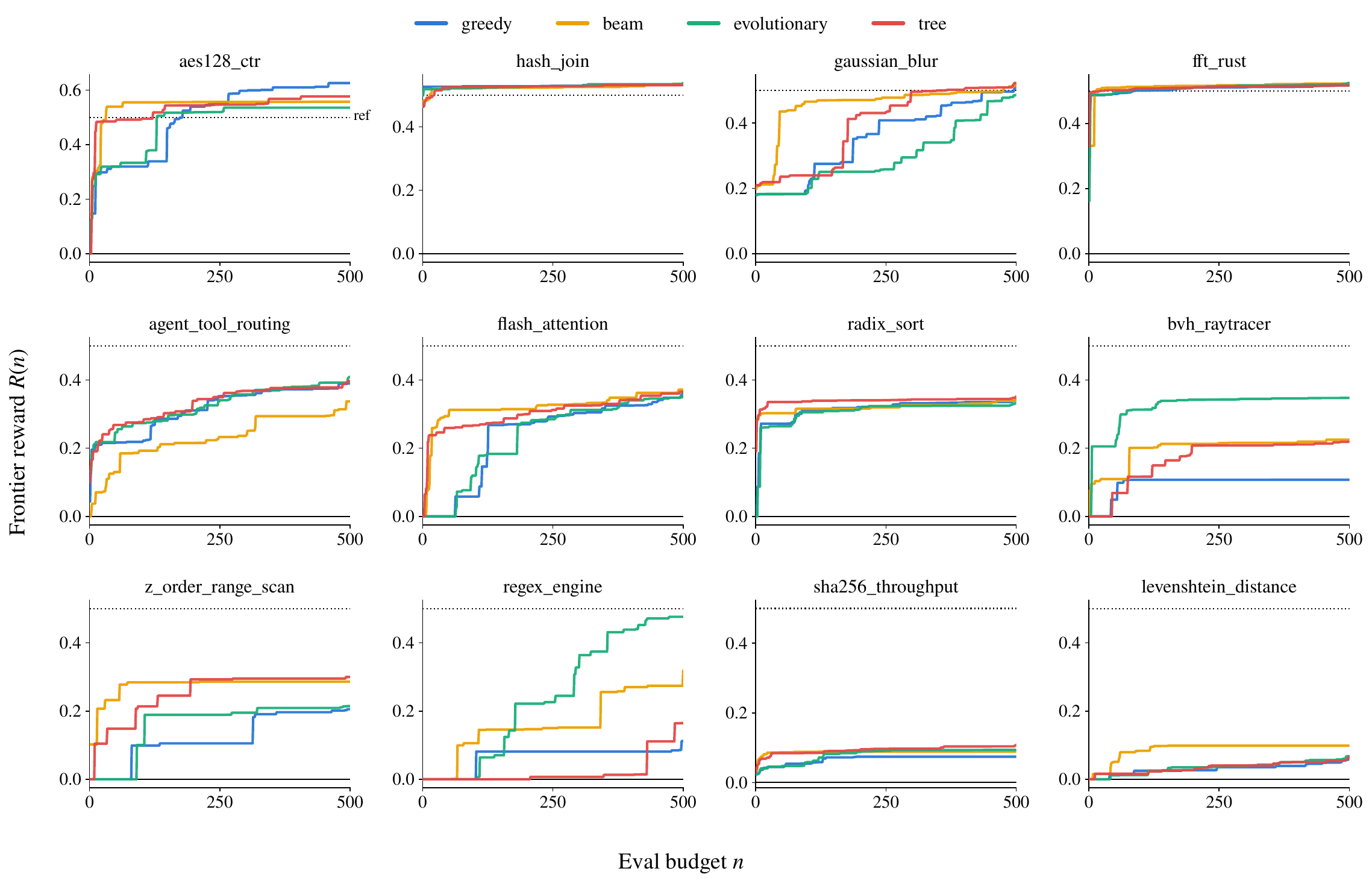}
    \caption{\scriptsize
    Frontier reward \(R(n)\) versus evaluation budget \(n\) for four fixed search policies across twelve AR tasks (mean over three seeds; the dotted line marks the reference level). Search strategy strongly shapes the Pareto frontier: policies that approach a similar endpoint can differ sharply in how quickly they reach it, and no single policy leads on every task.}
    \label{fig:trajectories}
\end{figure*}

\section{Research Efficiency}

\subsection{Research as a Search Process}

Search over a solution space is a natural paradigm for problem solving~\citep{russell2020artificial}. The research process implemented by most AR systems can likewise be viewed as a search procedure comprising a \textit{proposal policy} that generates candidate hypotheses or solutions, a \textit{verifier} that evaluates them, and a \textit{search structure} that records and organizes the states explored so far. This procedure repeats until the system finds a fully verified solution or exhausts its budget.

Methods such as Tree of Thoughts and Language Agent Tree Search make this structure explicit by expanding, evaluating, and selecting partial solutions according to a search policy~\cite{tot,lats}. More broadly, test-time computation is increasingly allocated to searching over candidate outputs against a verifier, rather than producing an answer in a single forward pass~\cite{snell2024scaling}. AR and reasoning differ primarily in the objects being searched and the criteria used to evaluate them: reasoning systems typically score partial reasoning states using a learned or self-assessed value function, whereas research systems often evaluate more complete artifacts using external task-specific verifiers. At an abstract level, however, both instantiate the same propose $\rightarrow$ score $\rightarrow$ select loop. This shared formulation allows tools from search, optimization, and sequential decision-making to be applied to the analysis of AR systems.

As language models make candidate generation abundant and cheap, verification rather than generation becomes the scarce resource: agents act as ``conjecture machines'' that produce hypotheses and solutions faster than these can be validated, and across most scientific disciplines the gap between proposing and confirming is widening~\cite{deepmind2026conjecture}. When verification is a cheap automatic check, the imbalance is mild; when verification is costly, the number of verifications a search can afford becomes the binding constraint. The count of verifications is therefore the natural budget for a research search, and how well a policy spends that budget, its efficiency, is the property we study.

\medskip\noindent\textbf{The Pareto frontier.}\enspace 
Over a run, an AR system proposes and evaluates a sequence of research artifacts (hypotheses, designs, programs)
\[
x_1, x_2, \ldots, x_N,
\]
each scored by a task verifier at some cost; let \(r_i\) be the reward of the \(i\)-th evaluated artifact. As argued above, and although a budget could equally be counted in tokens or wall-clock, we take it to be the number of evaluations \(n\): the unit is the trial itself, which isolates the search's intrinsic sample efficiency and keeps policies directly comparable at a fixed number of evaluations.

Because a research system keeps its best result, the object of interest is not the raw \(r_i\) but the frontier reward after \(n\) evaluations,
\[
R(n) = \max_{i \le n} r_i,
\]
a non-decreasing step function of the budget. This is the Pareto frontier of reward against the evaluation budget, the primitive from which every efficiency metric below is derived.

\subsection{Search Efficiency Matters}

AR systems are typically evaluated primarily by outcome quality: the final score, pass rate, or competition ranking achieved when the available budget is exhausted. Prominent benchmarks report this terminal performance~\cite{swebench, mlebench, kernelbench, autolab}, and studies comparing alternative search structures likewise rank systems by their final outcomes~\cite{toledo2025aira}. Even when the evaluation budget is varied, each budget setting is generally treated as an independent experiment producing a separate endpoint, rather than as a point on a budget-reward curve whose shape should itself be optimized.

Terminal performance alone, however, provides an incomplete measure of system quality. Search efficiency matters alongside final outcome. Given two systems that eventually achieve the same level of performance, if one reaches that level using only one-tenth of the computation, tool calls, experiments, or validation budget required by the other, it is clearly the more capable system under resource constraints. Recent work shows that, even with the same underlying model, execution time on the same task can differ by approximately 40\% depending on the design of the agent harness~\citep{ali2026harnesseffectorchestrationdesign}. The same principle applies to AR systems: orchestration, search policy, memory, verification, and stopping criteria can substantially affect the resources required to reach a given result. An outstanding AR system should produce not only high-quality solutions, but also find the solutions efficiently.

What is more, evaluating systems primarily by terminal outcome is natural when candidate verification is cheap. In coding or mathematics, for example, a candidate solution may be checked by running unit tests or evaluating a proof or formula, which are relatively inexpensive and fast. This assumption breaks down as AR expands into broader scientific domains in which verification is costly, such as wet-lab experimentation, prototype fabrication, clinical evaluation, or physical simulation. In these settings, every trial incurs a meaningful cost in time, money or materials. Reaching the same level of solution quality with fewer experiments may therefore determine whether an AR program is economically and operationally feasible.

Efficiency should consequently be treated as a first-class objective rather than as a secondary reporting metric. An AR system should be evaluated by the full relationship between expended budget and achieved reward, not only by the reward obtained at the largest available budget.

\section{Measuring Search Performance}

A search campaign produces a complete trace of intermediate rewards, not merely a final outcome. From this trace, we extract several complementary statistics that characterize both the quality and efficiency of a search run, as summarized in Table~\ref{tab:summaries}.

\begin{table}[h]
\centering
\small
\begin{tabular}{p{3.1cm} p{3.0cm} p{9.1cm}}
\toprule
Name & Summary & What it measures \\
\midrule
\textbf{best reward} & \(\hat{R} = R(N)\) & The \emph{outcome} axis: the peak reached by the end of the budget. \\
\textbf{AUC reward} & \(\bar{R} = \frac{1}{N}\sum_{n=1}^{N} R(n)\) & The \emph{efficiency} axis: the mean frontier reward over the budget (the normalized area under the frontier); of two runs that reach the same peak, the one that got there earlier scores higher. \\
\textbf{task ceiling} & \(R^{\star}\) & The best reward attained on the task by any policy and seed: an empirical measure of the task's optimizability, and the scale for cross-task normalization. \\
\textbf{convergence ratio} & \(\bar{R}/\hat{R} \in [0,1]\) & \emph{Speed} isolated from \emph{height}: the fraction of its final height the run holds over the budget (\(1\) means the peak was reached instantly). \\
\bottomrule
\end{tabular}
\caption{\scriptsize The scalar summaries of the Pareto frontier \(R(n)\) over a budget of \(N\) evaluations.}
\label{tab:summaries}
\end{table}

The \emph{best reward} is the metric by which AR systems are most commonly evaluated. However, it is strongly influenced by the intrinsic optimizability of the task and the capability of the underlying LLM. It therefore provides only a weak signal for improving the efficiency of the search policy itself. In contrast, the \emph{AUC reward}, $\bar{R}$, rewards search policies that discover high-quality artifacts early. It captures the optimizer's \emph{anytime} performance~\cite{zilberstein1996anytime, gao2022pmo}: the quality of the best available result throughout the run, and thus the improvement obtained per costly evaluation, rather than only the value eventually reached. We therefore propose AUC reward as the primary optimization target for improving search-policy efficiency. The \emph{convergence ratio} relates AUC reward to best reward by measuring how quickly a search run approaches its final peak. An ideal system achieves \(\bar{R}/\hat{R}=1\), indicating that it identifies its best solution at the very start of the search.


\section{Search Strategy Strongly Shapes Discovery Dynamics}

\begin{figure*}[h]
    \centering
    \includegraphics[width=\linewidth]{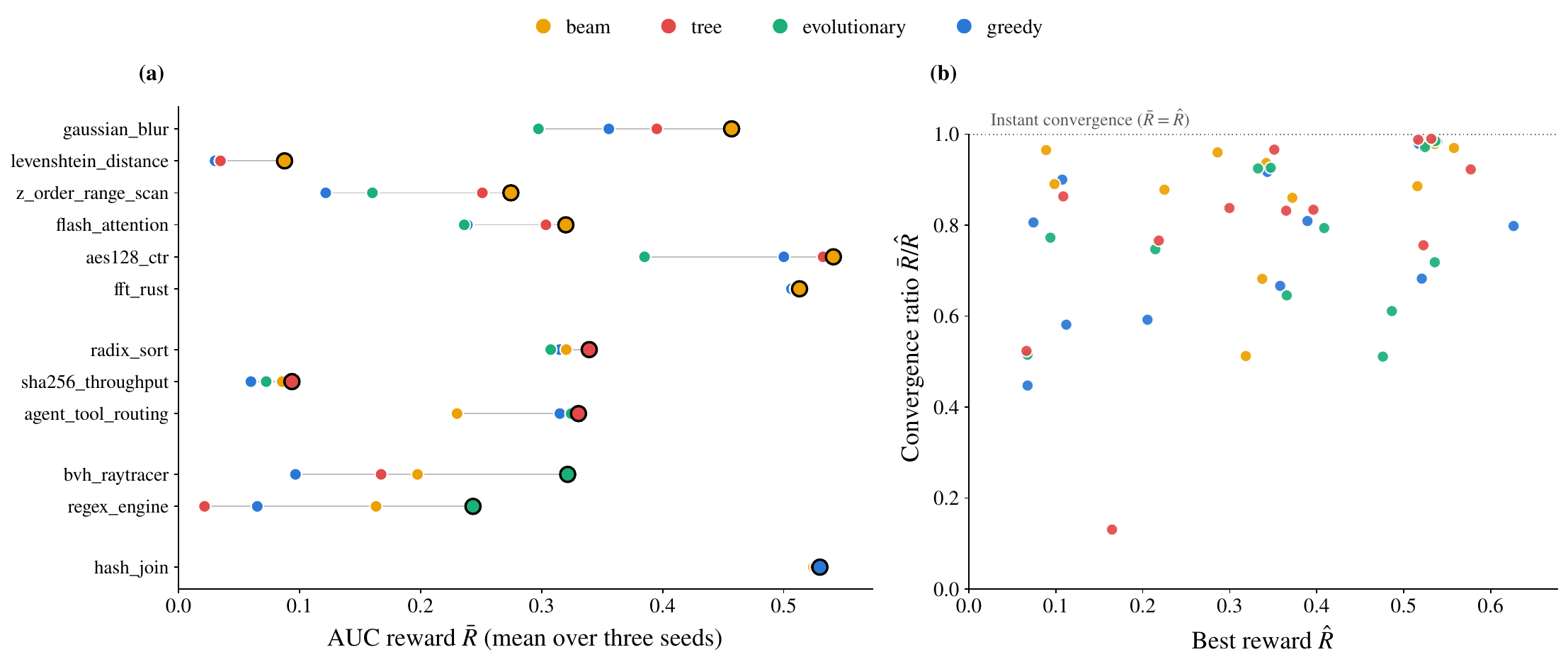}
    \caption{\scriptsize The four fixed policies across the twelve tasks (colors shared across panels). (a) AUC reward \(\bar{R}\) per task (mean over three seeds), rows grouped by the winning policy (black ring); overlapping dots are near-ties, and the long rows are the tasks where the choice of search structure matters most. (b) Best reward \(\hat{R}\) versus convergence ratio \(\bar{R}/\hat{R}\). The markers indicate the metrics per task and policy.}
    \label{fig:fixed-policies}
\end{figure*}

We evaluate on twelve systems-optimization tasks from AutoLab~\cite{autolab}, each a program whose measured performance is the research target.~\footnote{AutoLab's per-candidate reward is task-relative and log-scaled: \(r=0\) is the naive baseline and \(r=0.5\) matches the task's reference solution, gated to \(0\) for any candidate no faster than baseline.}
The proposer is Qwen3-Coder-480B (FP8). Each proposal is a single completion, conditioned on the task and the prior evaluation history, with no multi-turn agentic loop. 

We compare four fixed search policies (\texttt{greedy}, \texttt{beam}, \texttt{tree}, and \texttt{evolutionary}) under a matched frontier width of 64, defined as the number of live candidates maintained by each policy (parallel chains, beam entries, tree-expansion slots, or population members, respectively). Each policy is allocated a budget of 500 evaluations and run with three seeds per task. Final results are averaged across seeds. Motivated by the observed strengths and limitations of these fixed policies, we introduce the adaptive \texttt{fluid} search policy in Section~\ref{sec:fluid}.

Figure~\ref{fig:trajectories} plots the frontier reward \(R(n)\) for each fixed policy across all twelve tasks. Two patterns stand out.
First, policy performance diverges on difficult tasks but converges on easy, saturated ones. The curves cluster near the reference when little headroom remains (\texttt{hash\_join}, \texttt{fft\_rust}), but separate substantially when search strategy matters (\texttt{gaussian\_blur}, \texttt{z\_order\_range\_scan}, \texttt{regex\_engine}, \texttt{bvh\_raytracer}).
Second, different policies dominate at different budget regimes. \texttt{beam} is repeatedly the quickest starter: on \texttt{gaussian\_blur}, \texttt{flash\_attention}, and \texttt{levenshtein\_distance} it approaches its plateau while the other policies are still improving slowly. At larger budgets, however, \texttt{tree} and \texttt{evolutionary} often catch up or overtake it.

No search structure is universally most efficient, the best policy depends on the task landscape. Figure~\ref{fig:fixed-policies}a shows the winning policy for each task. The two clearest winners occupy opposite regimes: \texttt{beam} excels when rapid frontier deepening reaches the performance plateau early (\texttt{gaussian\_blur}, \texttt{flash\_attention}, \texttt{levenshtein\_distance}). In contrast, \texttt{evolutionary} performs best on crash-prone tasks (\texttt{bvh\_raytracer}, \texttt{regex\_engine}), where maintaining a population reduces the risk of catastrophic search failure. Furthermore, we decompose efficiency \(\bar{R}\) into outcome \(\hat{R}\) and convergence ratio \(\bar{R}/\hat{R}\), and the two factors are far from being collinear (Pearson \(r=0.40\), Figure~\ref{fig:fixed-policies}b).

Two implications follow. First, tasks with similar outcomes can have very different convergence profiles (Figure~\ref{fig:fixed-policies}b). Some reach a high reward only gradually (\texttt{regex\_engine}), whereas others reach a comparable level almost immediately (\texttt{hash\_join}, \texttt{fft\_rust}). Outcome alone cannot distinguish these cases, but efficiency can because it penalizes slow improvement. Second, outcome and efficiency can favor different policies. On \texttt{aes128\_ctr}, \texttt{greedy} reaches the highest outcome through prolonged refinement, while \texttt{beam} converges more quickly and is therefore more efficient. We also note that the finding that no search policy is universally superior is consistent with the argument in the recent work \cite{gupta2026automateddiscoveryuniversallysuperior}.

\section{\texttt{fluid}: A Budget-Adaptive Search Policy}
\label{sec:fluid}

\begin{figure}[h!]
    \centering
    \includegraphics[width=\linewidth]{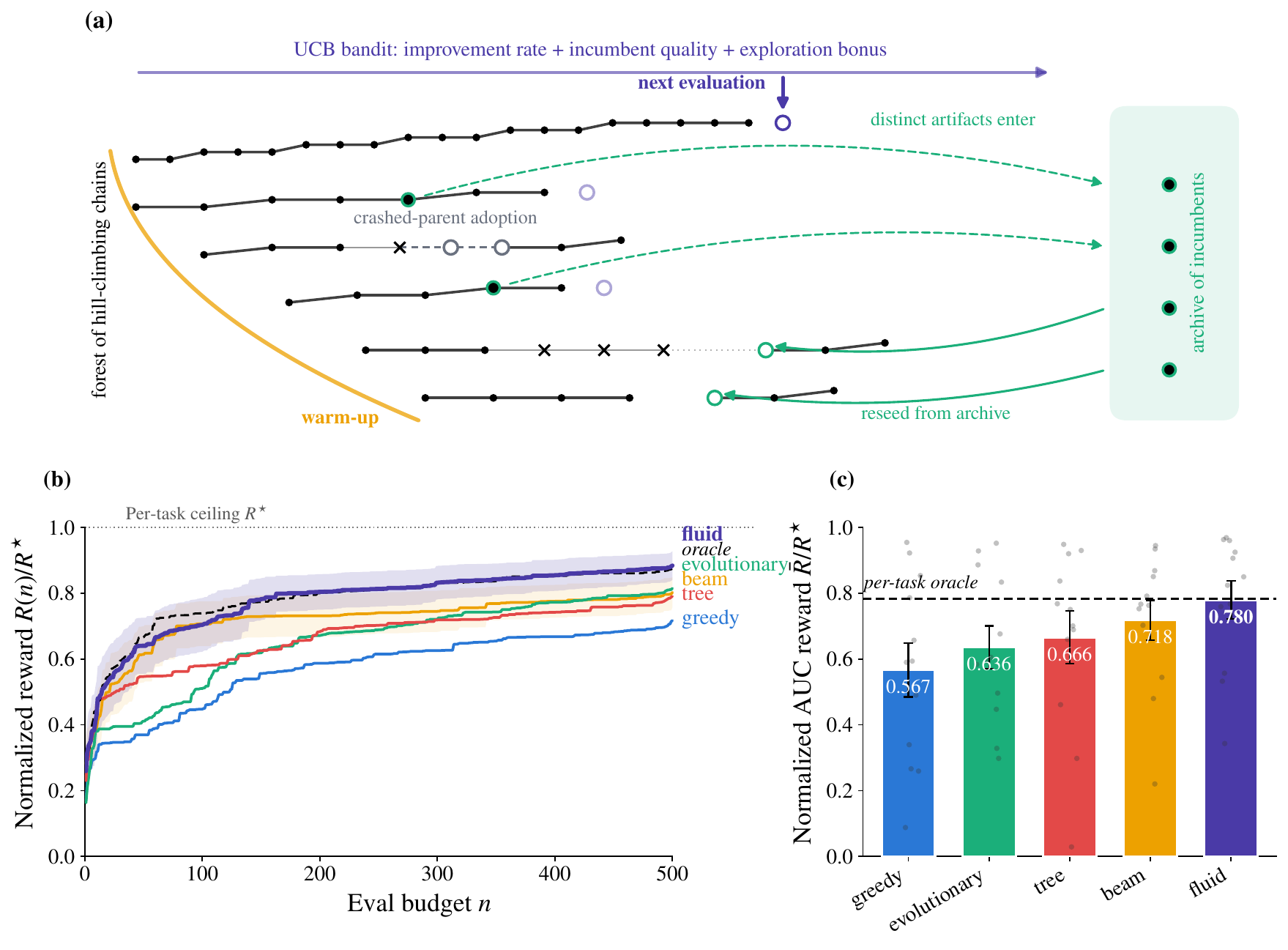}
    \caption{\scriptsize 
    The mechanism and efficiency of the proposed \texttt{fluid} search policy. \textbf{(a)}~Each row is a hill-climbing chain, a staircase rises when its incumbent improves (dots: evaluations; \(\times\): crashes). The arrow on top indicates the shared budget. At each step the UCB bandit fills one open slot (faint rings), so favored chains become denser while neglected ones stay sparse. Two levers are imported from fixed policies: orange represents the warm-start breadth ramp (\texttt{beam}'s early exploration, the forest widens as the pool opens \(16\to64\)), green represents \texttt{evolutionary}'s archive (filled rings store distinct artifacts, open rings reseed a stalled chain). Grey marks a \texttt{fluid}-native lever, a crashed-parent adoption that keeps editing a crashed artifact for a few steps rather than discarding it. \textbf{(b)}~Normalized frontier reward \(R(n)/R^{\star}\) averaged over the twelve tasks, bands indicate \(\pm 1\) standard error on \texttt{fluid} and \texttt{beam}, dashed curve indicates the per-task oracle. \textbf{(c)}~Normalized AUC reward \(\bar{R}/R^{\star}\) per policy (bars indicate the mean over tasks, error bars \(\pm 1\) standard error, dots indicate per-task metrics). It can be seen that \texttt{fluid} is the most efficient single policy and essentially reaches the oracle.}
    \label{fig:fluid}
\end{figure}

The preceding analysis indicates that efficiency is an optimization objective distinct from the final outcome, and that no fixed search structure is uniformly most efficient. The best policy is task-dependent. Viewed through a common lens, each fixed structure is a predetermined rule for allocating the evaluation budget. \texttt{greedy} distributes evaluations evenly across independent chains, each refining its own incumbent; \texttt{beam} concentrates each round on the current top candidates and discards the rest; \texttt{tree} directs evaluations toward promising branches of a single search tree; and \texttt{evolutionary} allocates them across a maintained population, selecting parents by tournament. Because the best allocation pattern is task-dependent, it cannot be chosen reliably before the run. It must instead be inferred online from evidence generated during the run itself.

We therefore propose \texttt{fluid} search, a budget-adaptive portfolio bandit over a forest of hill-climbing chains. The high-level idea is depicted in Figure~\ref{fig:fluid}a, and detailed algorithmic steps are given in Appendix~\ref{Appendix:fluid_algo}. The method brings bandit allocation for multi-start local search~\cite{multistart, ils, rice1976algselection, gomes2001portfolios} to LLM-driven AR, where each edit is proposed by a model conditioned on the run history rather than by a fixed mutation kernel. Each chain maintains one incumbent artifact and proposes edits from it, and a UCB-style rule~\cite{auer2002ucb} allocates the next evaluation across chains by scoring chain \(k\) as
\[
U_k \;=\; \rho_k \;+\; w\,q_k \;+\; c\,\sqrt{\frac{\ln (n+1)}{n_k+1}},
\]
where \(\rho_k\) is the fraction of chain \(k\)'s five most recent evaluations that improved its incumbent, initialized to \(\tfrac{1}{2}\) for a chain that has not yet been evaluated, \(q_k \in [0,1]\) its incumbent reward normalized to the best artifact found so far, \(n_k\) the number of evaluations it has received, and \(n = \sum_k n_k\) the total so far. We fix \(w = 1\) and \(c = \tfrac{1}{2}\).

The first two terms are the exploitation score: \(\rho_k\) favors chains that are still making progress, while \(q_k\) favors chains already near the current frontier. The third term is the standard exploration bonus, which is largest for under-sampled chains. The resulting allocation concentrates budget on productive directions without allowing any chain to starve. Two design conventions are important. First, until some artifact exceeds the baseline, we set \(q_k=0\) for all chains. Allocation in this regime therefore depends only on recent improvement and exploration, which is also the regime targeted by the crash-repair mechanism described below. Second, when a chain is restarted, its improvement-rate estimate is reset to the unevaluated prior, \(\rho_k=\tfrac{1}{2}\). The restarted chain is thus judged on fresh evidence rather than penalized by the poor incumbent that triggered the restart.

At each step, the rule selects the highest-scoring \emph{idle} chain, meaning one that is not already awaiting a verifier result, because evaluations run concurrently.~\footnote{Evaluations are asynchronous and decoupled from proposal generation. The harness keeps up to 32 proposal-and-verification jobs in flight across 64 chains. While some chains await the compile-run-measure verifier, newly available slots are assigned to the highest-scoring idle chains. During the warm-start ramp described below, the in-flight cap is 8, preserving the same one-half ratio between active jobs and live chains. Maintaining this ratio leaves idle chains available for bandit selection. Concurrency affects wall-clock throughput but not the evaluation budget.} Building on this core loop, \texttt{fluid} carries five independent levers, each motivated by the fixed-policy results. With all of them disabled, it reduces to the bare bandit forest above, making the contribution of each mechanism separately identifiable. The bandit determines which chain receives the next evaluation, and the mechanisms reduce the chance that the allocated evaluation is wasted.

\begin{itemize}

\item A run-global archive of distinct incumbents, combined with sampled reseeding, imports the population diversity that allowed \texttt{evolutionary} search to withstand crash-prone tasks. When a chain is restarted, it can resume from a previously discovered artifact rather than always returning to the baseline.

\item A reseeding rule abandons chains whose proposals repeatedly crash or whose incumbents have ceased to improve. Complementing it, bounded crashed-parent adoption allows a chain to continue editing its own crashed artifact for a small number of steps, with the verifier error included in context. This gives locally repairable failures a chance to be fixed rather than forcing the search to rediscover them from scratch.

\item A merge-softening rule suppresses redundant work. When two idle chains converge to the same artifact, one is reseeded from the archive, but only if the duplication persists across several checks. Chains holding the baseline are exempt. These safeguards prevent transient agreement, or the temporary synchronization of newly reseeded chains, from collapsing population diversity prematurely.

\item A warm-start breadth ramp concentrates the opening budget on a smaller pool of chains, allowing each lineage to deepen more rapidly. This imports the early depth compounding that made \texttt{beam} the fastest-starting fixed policy.~\footnote{\texttt{beam}'s fast start is structural. The loop is standard beam search, expanding every node of the current layer into three children and keeping at most \(64\) of the pooled children, but the first layer is the baseline alone, so the layers grow \(1, 3, 9, 27\) and pruning first binds when \(81\) children exceed the cap. The opening rounds are therefore cheap: within roughly the first \(120\) evaluations \texttt{beam} stacks four generations of edits, where a policy running \(64\) parallel chains spends the same budget on one or two proposals per chain. The ramp imitates this geometric opening by concentrating the early budget on \(16\) chains.}
\end{itemize}

\noindent \texttt{fluid} is evaluated on the same tasks, random seeds, and \(500\)-evaluation budget as the fixed policies. It maintains a maximum pool of \(64\) chains, matching their frontier width, while the warm-start ramp restricts the active pool to \(16\) chains through evaluation \(150\).

After normalizing each task's reward by its observed ceiling \(R^{\star}\) and averaging across all twelve tasks (Figure~\ref{fig:fluid}c), \texttt{fluid} emerges as the most efficient single policy in the suite. It achieves a normalized AUC reward of \(0.780\), outperforming \texttt{beam}, the strongest fixed policy at \(0.718\), and essentially matching the per-task efficiency oracle at \(0.784\). 
The oracle selects, for each task, the fixed policy with the highest mean normalized AUC on the same runs and thus represents performance under advance knowledge of the best search structure. \texttt{fluid} already matches \texttt{beam}, the fastest fixed-policy starter, over the opening budget, then steadily widens its advantage through the remainder of the run (Figure~\ref{fig:fluid}b).

\section{Discussion}

Discovery efficiency, improvement per costly evaluation, should be treated as a first-class objective for AR. This perspective changes both how these systems are evaluated and how they should be designed. Our results show that search efficiency and final outcome quality are empirically distinct dimensions. Across the classical search structures we study --- hill climbing, tree search, beam search, and evolutionary search --- no fixed structure is universally the most efficient. This heterogeneity motivates adaptive policies that allocate search budget dynamically according to the demands of each task. Based on this observation, we introduce \texttt{fluid}, a budget-adaptive policy that nearly matches a per-task oracle without prior knowledge of which search structure is best suited to the task.

The broader design principle supported by our results is \emph{budget adaptivity}. A fixed search structure commits in advance to a particular way of spending the evaluation budget. Yet the properties that determine which allocation is appropriate—such as the task's crash rate, the reliability of local refinement, and whether early gains compound—become observable only through the evaluations performed during the run. Budget allocation should therefore be treated as a sequential decision, continually revised using evidence from the emerging search trajectory. \texttt{fluid} provides one instantiation of this principle, and its near-oracle efficiency suggests that the trajectory reveals useful allocation signals early enough for the system to act on them. This principle is especially consequential in domains where each evaluation corresponds to an expensive experiment and the remaining research budget must be redirected as evidence accumulates. In its current form, however, \texttt{fluid} adapts only the allocation among a predefined set of hill-climbing chains. A natural extension is a heterogeneous portfolio whose members implement different search structures, allowing the system to adapt not only the distribution of resources but also the form of search itself.

In the ladder of innovation autonomy proposed by Cao and Yang~\cite{cao2026beyond}, our setting corresponds to the first autonomous level: a closed loop that improves a candidate within a fixed representational frame and evaluates it using a fixed verifier. This is the regime exemplified by systems such as FunSearch~\cite{funsearch} and AlphaEvolve~\cite{alphaevolve}. Our contribution is to make the search-efficiency dimension of this regime explicit.
Efficient search within a fixed frame, however, is necessary but not sufficient for more general research autonomy. On several tasks, even the strongest policies eventually plateau. One possible explanation is that these plateaus reflect a limitation of the proposer: the model may be unable to express or reliably realize the representational primitive required for further progress. This constitutes a form of the ``vocabulary gap''—the difficulty of inventing and stabilizing a new representational primitive rather than merely recombining primitives already available. More efficient traversal of a fixed search space cannot recover a solution that lies outside the space induced by the proposer. A second limitation arises from the verifier. In our experiments, verification is fixed, inexpensive, formal, and immediate. This controlled setting allows us to isolate the effect of search structure, but it excludes discoveries whose value becomes apparent only through subsequent reuse or interaction with later developments. Such cases exhibit a ``verifier gap'': a potentially valuable change cannot be recognized by the current evaluation procedure because the evidence required to assess it does not yet exist.

The distinction between efficiency and outcome can therefore be understood as the within-frame instance of a more general distinction. Search efficiency measures how rapidly a system makes progress within a fixed space; it does not determine whether that space is itself adequate. A policy may be maximally efficient yet remain constrained by a representational ceiling, just as another policy may eventually reach a strong solution while converging too slowly to be practical. Efficiency is therefore the appropriate objective for search within a fixed frame, but it does not by itself capture transformations that expand, revise, or replace the frame.

Two natural extensions connect the present setting to higher levels of research autonomy. The first is \emph{generative editing}: extending the proposer from modifying a single artifact within a fixed vocabulary to inventing reusable primitives that reshape the space of future solutions. The second is \emph{adaptive verification}: allowing evaluation procedures to co-evolve with the representation and to recognize discoveries whose value is not visible under the original metric.

These extensions do not diminish the importance of efficiency. Systems that invent new primitives or revise their own verifiers still consume computational, experimental, and human resources. Their performance curves may be more complex than the frontier reward studied here, but the underlying question remains unchanged: how much useful progress is produced for a given budget? The adaptive principle demonstrated in this work—maintaining multiple search processes, avoiding premature commitment to a single structure, and reallocating resources to processes that continue to yield progress—provides a natural foundation for addressing this question at higher levels of research autonomy.

\bibliographystyle{unsrtnat}
\bibliography{references}

\clearpage
\appendix

\section{The \texttt{fluid} Search Algorithm}
\label{Appendix:fluid_algo}

\begin{algorithm}
\caption{\texttt{fluid}: Budget-Adaptive Portfolio Search}
\small
\begin{algorithmic}[1]
\Require Baseline $x_0$, budget $B$, chain counts $M_0 \le M$,
warm-start threshold $T_w$
\Ensure Best valid artifact found

\State Initialize $M$ chains with $x_i \gets x_0$
\State Initialize archive $\mathcal{A} \gets \{x_0\}$ and $N \gets 0$

\While{$N < B$}
    \State $\mathcal{I} \gets
    \{1,\ldots,M_0\}$ if $N<T_w$; otherwise
    $\{1,\ldots,M\}$

    \State Select the highest-priority idle chain
    \[
        i \gets
        \arg\max_{j\in\mathcal{I}:\,\mathrm{idle}}
        \left[
        \rho_j+wq_j+
        c\sqrt{\frac{\ln(N+1)}{n_j+1}}
        \right]
    \]

    \State $y \gets \Call{ProposeEdit}{x_i}$; evaluate $y$
    \State $N \gets N+1$; $n_i \gets n_i+1$; update $\rho_i$

    \If{$y$ is valid}
        \State $\Call{ArchiveUpdate}{\mathcal{A},y}$; reset crash count
        \If{$f(y)>f(x_i)$}
            \State $x_i\gets y$; reset stagnation count
        \Else
            \State increment stagnation count
        \EndIf
    \Else
        \State increment crash count
        \If{crash repair is allowed}
            \State $x_i\gets y$; retain verifier error as context
        \EndIf
    \EndIf

    \If{$i$ is non-best and repeatedly crashes or stagnates}
        \State $\Call{Reseed}{i,\mathcal{A},x_0}$
    \EndIf

    \ForAll{persistent duplicate pairs $j<k$ with
            $x_j=x_k\neq x_0$}
        \State \textsc{Reseed} one non-best chain in $\{j,k\}$
    \EndFor
\EndWhile

\State \Return the highest-scoring valid artifact found
\end{algorithmic}
\end{algorithm}

Here, $f$ denotes the verifier score ($-\infty$ for an invalid artifact, so any valid candidate replaces a crashed incumbent and ends its repair), $\rho_i$ is the fraction of the most recent evaluations of chain $i$ that improved its incumbent, $q_i$ is the incumbent reward normalized by the best reward found so far, $n_i$ is the number of evaluations allocated to the chain, and $w$ and $c$ are the allocation weights of Section~\ref{sec:fluid}. Unevaluated and reseeded chains use the prior $\rho_i=1/2$.

\textsc{ArchiveUpdate} retains diverse valid artifacts, replacing a similar archived artifact only when the new artifact scores higher. When the archive exceeds its capacity, its lowest-scoring replaceable entry is removed. \textsc{Reseed} restarts a chain from the baseline with a small fixed probability $\epsilon$ and otherwise from a strong, underused archive entry; all chain-local statistics are then reset.

\end{document}